\documentclass{article}

 \usepackage[preprint,nonatbib]{neurips_2025}
 
\usepackage{amsmath,amssymb,amsthm,amsfonts}

\usepackage[pdftex]{graphicx}

\usepackage{booktabs}
\usepackage{subcaption}

\usepackage[utf8]{inputenc} 
\usepackage[T1]{fontenc}    
\usepackage{hyperref}       
\usepackage{url}            
\usepackage{booktabs}   
\usepackage{algorithm}
\usepackage{amsfonts}       
\usepackage{nicefrac}       
\usepackage{microtype}      
\usepackage{xcolor}         

\title{Moment Estimates and DeepRitz Methods on Learning Diffusion Systems with Non-gradient Drifts }

\author{%
  Fanze Kong\thanks{Corresponding author}  \\
  Department of Applied Mathematics\\
  University of Washington\\
  Seattle, WA 98105, USA\\
  \texttt{fzkong@uw.edu} \\
  \And
  Chen-Chih Lai \\
  Department of Mathematics\\
  Columbia University\\
  New York, NY 10027, USA\\
\texttt{cclai.math@gmail.com} \\
  \AND
  Yubin Lu \\
  Department of Applied Mathematics \\
  Illinois Institute of Technology \\
  Chicago, IL 60616, USA\\
   \texttt{ylu117@illinoistech.edu} \\
}

\begin{document}

\maketitle

\begin{abstract}
Conservative-dissipative dynamics are ubiquitous across a variety of complex open systems. We propose a data-driven two-phase method, the Moment-DeepRitz Method, for learning drift decompositions in generalized diffusion systems involving conservative-dissipative dynamics. The method is robust to noisy data, adaptable to rough potentials and oscillatory rotations. We demonstrate its effectiveness through several numerical experiments.
\end{abstract}

\section{Introduction}
Nonlinear stochastic dynamics are ubiquitous in complex systems across physics, biology, and engineering.  Incorporating the randomness perturbations (e.g.  {Gaussian} noise) in deterministic ordinary differential equations (ODEs), stochastic differential equations (SDEs) provide a powerful mathematical framework for modeling such complex dynamics \cite{Evansbook,sarkka2019applied}.  By employing simple SDEs with gradient drifts, whose density functions are governed by potential Fokker-Planck equations (typically Langevin equations), a huge volume of research has been done for investigating the  equilibrium behaviors of particle systems including the relation between temperature, pressure and entropy at the macroscopic level \cite{sekimoto1998langevin,hatano2001steady,quigley2004langevin}.  However, while investigating the non-equilibrium thermodynamics of particle systems, various complex phenomena occur, including entropy production \cite{esposito2012stochastic,crooks1999entropy,qian1991reversibility}, conservative-dissipative dynamics \cite{kurchan1998fluctuation,marconi2008fluctuation,esposito2010three}, non-Boltzmann steady states \cite{dorfman1999introduction,jiang2004mathematical}, etc.  Non-detailed balance is a crucial ingredient in the occurrence of intricate statistical behaviors of collective particles mentioned above and FP equations with the non-gradient structures serve as paradigms to understand such property. 

Research on the discovery of physical laws described by ODEs and SDEs from observational data via data-driven approaches has been growing rapidly, see e.g.  \cite{brunton2016discovering,raissi2019physics,weakSINDy4PDE,huang2022variational,variationalPINNs,lu2024learning}.  The main idea is based on the machine learning framework, for example neural networks are trained to minimize suitable {physics-informed} loss functions.  The loss functions can take various forms, such as PDE-based forms {\cite{brunton2016discovering,raissi2019physics,PINNinverse2}}, {weak} forms \cite{Lu_self-test,zang2020weak} and {variational} forms {\cite{MLforIrreversibleProcess1,stat-PINNs,lu2024learning,KLL2025}}.  Due to the fundamental connection between non-detailed balance and non-gradient drift structures, the precise identification of drift decomposition plays a crucial role in the investigation of  non-equilibrium thermodynamics.  On one hand, the rotational components of the drift provide a way to compute the entropy production of the diffusion process evolving on Riemannian manifolds with non-trivial topology \cite{min1999entropy}.  On the other hand, the computation of irrotational components is significant in the large deviation theory \cite{lin2022data}.   Motivated by the results of Lu et al. \cite{lu2024learning}, Kong at al. \cite{KLL2025} proposed a two-stage data-driven framework for learning the drift decomposition, satisfying pointwise orthogonality, in generalized diffusions by combining the evolution of first moments with the energy dissipation law.

 In this paper, our objective is to extend the results in \cite{KLL2025} and recover the drift decomposition in FP equations with the more general non-gradient drifts.  We remark that, Kong et al.  \cite{KLL2025} only address the case in which the drift terms satisfy a pointwise orthogonal decomposition, whereas the present work proposes a data-driven framework for tackling  general scenarios by combining the first-moment dynamics with DeepRitz method.

In Section \ref{framework}, we introduce the Moment-DeepRitz method 
for identifying pseudo-potential and rotation components in diffusion systems. Section \ref{examples} presents two-dimensional experiments, and Section \ref{conclusions} summarizes results and open problems.

\section{Learning framework}
\label{framework}

We present the Moment-DeepRitz method for learning generalized diffusions with non-gradient structures, starting from the stochastic system governed by the following SDE:
\begin{align}\label{SDE1}
d{\bf X}_t = {{\bf b}({\bf X}_t)}dt + \sigma({{\bf X}_{t}}) d{\bf W}_t,~X_t=(x^t_1,\cdots,x^t_d)\in \mathbb R^d,
\end{align}
where ${\bf b} = (b_1,\ldots,b_d): \mathbb R^d\rightarrow \mathbb R^d$ is a drift term and ${\bf W}_t$ denotes a standard $d$-dimensional Brownian motion with a scalar noise intensity $\sigma = \sigma({\bf X}_t)$.
The probability density function $f({\bf x},t)$ of ${\bf X}_t$ evolves according to the Fokker-Planck equation
\begin{align}\label{FPeq}
\left\{\begin{array}{ll}\partial_t f+\nabla\cdot({\bf b}f)=\frac{1}{2}\Delta (\sigma^2 f),&{\bf x}\in\mathbb R^d,t>0,\\
f({\bf{x}},0)=f_0,&{\bf x}\in\mathbb R^d,
\end{array}
\right.
\end{align}
where the classical It\^{o} integral formula has been applied to (\ref{SDE1}) and $f_0$ denotes the density of initial state ${\bf X}_0$.
The evolution of the first moment $\int_{\mathbb R^d} f({\bf x}){\bf x}\, d{\bf x}$ is given by (see, e.g., \cite{KLL2025})
\begin{align}\label{firstmomentdynamics}
\frac{d}{dt}\int_{\mathbb R^d}f {{x_i}}  d{\bf{x}}=\int_{\mathbb R^d} b_i f\, d{\bf x},\quad {\bf x} = (x_1,\ldots,x_d).
\end{align}
To further extract the potential and rotation components of the drift in generalized diffusions, we assume the drift $\bf {b}$ can be decomposed uniquely as
\begin{align}\label{driftdecomposition}
{\bf b} = -\nabla\psi + {\bf R},
\end{align}
where $\nabla\cdot{\bf R} = 0.$  Then, taking the divergence of both sides in (\ref{driftdecomposition}) yields 
\begin{align}\label{poisson}
    -\Delta\psi = \nabla\cdot{\bf b}.
\end{align}
Since $\psi$ solves the Poisson equation (\ref{poisson}), it is a minimizer of the following optimization problem
\begin{align}\label{variationalform}
\min\limits_{\psi\in H} I(\psi),~~I(\psi) 
&:= \frac12 \int_{\Omega} |\nabla\psi|^2\, d{\bf x} + \int_{\Omega}  \nabla\psi\cdot{\bf b}\, d{\bf x},
\end{align}
where $H$ is the set of admissible functions.

Combining (\ref{firstmomentdynamics}) with (\ref{variationalform}), we   propose a deep learning based method, the Moment-DeepRitz method for identifying the physical laws of generalized diffusions: in the first phase, to learn the general drift ${\bf b}$, one formulates the loss function  based on the first-moment dynamics (\ref{firstmomentdynamics}); while in the second phase,  one follows DeepRitz method in \cite{EBdeepritz2018} and employs the variational form (\ref{variationalform}) to construct the loss function and recover the irrotational component.

Given the noise intensity $\sigma^2$, we aim to implement the Moment-DeepRitz method in the continuous data observation setting. 
Here, we define ${\bf b}_{\rm NN}(x;
\theta)$ and $\psi_{\rm NN}(x;\theta)$ as neural network approximations of the drift and potential components ${\bf b}$ and $\psi(x)$, respectively.  

We construct the training data as follows. For $j=1,\ldots,M$, Gaussian-type initial states $(f_0)_j({\bf x})$ with mean $\boldsymbol{\mu}_j^0$ and variance $\sigma_0^2$ generate numerical solutions $f_j({\bf x},t)$ of  the FP equation (\ref{FPeq}). Let $\Delta x_k$ and $\Delta t$ be the spatial and temporal mesh sizes, and define observation grids $\delta x_k=\Delta x_k$, $\delta t=m\Delta t$. With uniform mesh points ${{\bf x}_i}$ of spacing $\delta{\bf x}=(\delta x_1,\ldots,\delta x_d)$, the training dataset is $\{f_j({\bf x}_i,t_1), f_j({\bf x}_i,t_2)\}_{i,j=1}^{N,M}$, where $t_2=t_1+\delta t$ and $t_1\ll1$ is a short transient time.

We now present the Moment-DeepRitz method (see Algorithm ~\ref{Alg:EnVarA} in Appendix~\ref{Alg:EnVarA}):

\noindent{\bf Phase 1: First-moment estimates.}
This phase is identical to Stage 1 of \cite{KLL2025}.
The objective is to solve the minimization problem 
\begin{equation}\label{eq-theta_b*}
\theta_{\bf b}^* = \underset{\theta}{\text{argmin}}\ L^{\rm dyn}_{\bf b}(\theta).
\end{equation}
where $L^{\rm dyn}_{\bf b}(\theta) = \sum_{k=1}^d L^{\rm dyn}_{b_k}(\theta)$ in which 
\begin{align}\label{Ldynamic}
L^{\rm dyn}_{b_k}(\theta) = \sum_{j=1}^M \left\|\frac{(\mu_j)_k(t_2) - (\mu_j)_k(t_1)}{t_2-t_1} - |\delta{\bf x}|\sum_{i=1}^N (b_k)_{\rm NN}({\bf x}_i;\theta) f_j({\bf x}_i, t_1) \right\|^2,\
k = 1,\ldots, d.
\end{align}
where
$(\mu_j)_k(t) = |\delta{\bf x}| \sum_{i=1}^N (x_i)_k f_j({\bf x}_i, t)$, ${\bf x}_i = ((x_i)_1,\ldots,(x_i)_d)$,
denotes the approximation of the centroid of density function $f$ solving (\ref{FPeq}) for $j=1,\ldots,M,$ and $k=1,\ldots,d $.



\noindent{\bf Phase 2: DeepRitz method.}
Approximating (\ref{variationalform}) by a Riemann sum yields the loss function
\begin{align}\label{idynloss}
I_{\psi}^{\rm dyn}(\theta) = |\delta{\bf x}| \sum_{i=1}^N \left[\frac{1}{2}|\nabla\psi_{\rm NN}(
{\bf x}_i;\theta)|^2+\nabla\psi_{\rm NN}({\bf x}_i;\theta)\cdot {\bf b}^*({\bf x}_i)\right],\quad 
{\bf b}^*({\bf x}_i):={\bf b}_{\rm NN}({\bf x}_i; \theta_{\bf b}^*)
\end{align}
where $\theta_{\bf b}^*$ is given by (\ref{eq-theta_b*}) in Phase 1.
Learning $\psi$ reduces to the optimization problem:
\begin{equation}\label{eq-psi-minimization}
\theta_\psi^* = \underset{\theta}{\text{argmin}}\ I_{\psi}^{\rm dyn}(\theta).
\end{equation}


\section{Numerical examples}
\label{examples}
We present representative two-dimensional experiments ($d=2$). Since the FP equation (\ref{FPeq}) is posed on $\mathbb{R}^2$, we approximate its solution on the domain $[-4,4]\times[-4,4]$. In Phase 1, observation data are generated from $M$ initial distributions $\mathcal{N}(\boldsymbol{\mu}^0,\sigma^2_0 I_2)$, where $\boldsymbol{\mu}^0$ is uniformly sampled from $[-2,2]\times[-2,2]$, $\sigma^2_0=0.01$, and $I_2$ denotes the $2\times2$ identity matrix.

The training dataset $\{f_j(x_i,y_i,t_1), f_j(x_i,y_i,t_2)\}_{i,j=1}^{N,M}$ is generated by numerically solving the FP equation with spatial resolution $\Delta x = \Delta y = 0.1$ and time step $\Delta t = 0.0001$. Unless otherwise specified, we construct $M=40$ distinct initial distributions and extract solution snapshots at $t_1 = 0.015$ and $t_2 = 0.016$ (i.e., $m=10$) for training.

To test the robustness of our method to noisy data, the training samples are perturbed with Gaussian noise of level $0.1$, obtained by convolving the numerical solution $f$ with the distribution $\mathcal{N}(\mathbf{0},\, 0.1 I_2)$.

For training in both phases, we employ a fully connected neural network with four layers: two hidden layers, each containing 50 neurons. The activation function is {\bf Tanh()}, and the parameters are optimized using the Adam algorithm with a learning rate of $1\times10^{-4}$. 
In Phase 1, the network is trained with a batch size of 5. 
Unless otherwise specified, both phases are trained for 10,000 epochs.

Since $\psi$ is defined up to an additive constant, we shift $\psi_{\rm NN}$ so its mean matches that of $\psi$, i.e.,
$\psi_{\rm NN}({\bf x}) \mapsto \psi_{\rm NN}({\bf x}) + \tfrac1N\sum_{i=1}^N (\psi({\bf x}_i)-\psi_{\rm NN}({\bf x}_i))$.

To assess the accuracy of our method at each phase, {we compute the relative root mean square error (rRMSE) of the drift ${\bf b}$ as follows: 
${\rm rRMSE}_{\bf b} = \sqrt{\sum_{i=1}^N {|{\bf b}({\bf x}_i) - {\bf b_{\rm NN}}({\bf x}_i; \theta_{\bf b}^*)|^2}}/\sqrt{\sum_{i=1}^N |{\bf b}({{\bf x}}_i)|^2}$. 
The rRMSE of $\psi$ (${\rm rRMSE}_\psi$) and ${\bf R}$ (${\rm rRMSE}_{\bf R}$) are defined in a similar fashion.}


\noindent\textbf{Double-well Potential.} 
Consider the potential $\psi(x,y) = \frac14(x^2-1)^2 + \frac12y^2$ and the rotation ${\bf R}(x,y) = [y,-x]^\top$, with the noise intensity $\sigma^2 = 2, 10^{-3},$ and $0$.

Using the Moment-DeepRitz method, we recover both the drift vector field and the associated pseudo-potential, which together yield the corresponding rotation field. 
The relative root mean square errors are reported in Table~\ref{tab:double-well}, and the first three rows of Figure~\ref{fig:double-well} compare the learned results with the ground truth.

\vspace{-0.45cm}
\begin{table}[H]
  \centering
  \begin{subtable}[t]{0.48\linewidth}
    \centering
    \caption{Learning of the double-well potential under different noise intensities.}
    \label{tab:double-well}
    \begin{tabular}{llll}
      \toprule
      $\sigma^2$ & ${\rm rRMSE}_{\bf b}$ & ${\rm rRMSE}_{\psi}$ & ${\rm rRMSE}_{\bf R}$ \\
      \midrule
      2      & 2.83e-02 & 7.27e-02 & 9.03e-02 \\
      $10^{-3}$  & 2.58e-02 & 8.07e-02 & 9.10e-02 \\
      0      & 2.50e-02 & 1.09e-01 & 1.09e-01 \\
      \bottomrule
    \end{tabular}
  \end{subtable}
  \hfill
  \begin{subtable}[t]{0.48\linewidth}
    \centering
    \caption{Learning of different potentials and rotations.}
    \label{tab:other-examples}
    \begin{tabular}{llll}
      \toprule
      Example & ${\rm rRMSE}_{\bf b}$ & ${\rm rRMSE}_{\psi}$ & ${\rm rRMSE}_{\bf R}$ \\
      \midrule
      QW $\psi$    & 1.37e-02 & 2.41e-01 & 7.07e-02 \\
      Osc. ${\bf R}$  & 6.44e-03 & 1.84e-01 & 2.17e-01 \\
      Rough $\psi$   & 2.64e-02 & 1.17e-01 & 1.33e-01 \\
      \bottomrule
    \end{tabular}
  \end{subtable}
  \vspace{0.2cm}
  \caption{Relative root mean square errors across different examples.}
  \label{table1}
\end{table}

\vspace{-0.75cm}

\noindent\textbf{Quadruple-well (QW) Potential.} 
Consider the potential $\psi(x,y) = \frac18(x^2-1)^2 + \frac18(y^2-1)^2$ and the rotation ${\bf R}(x,y) = [y,-x]^\top$, with the noise intensity $\sigma^2 = 2$.
We choose a larger number of datasets $M=80$ and increase the number of epochs to 100,000.

\noindent\textbf{Oscillatory Rotation (Osc. R).} 
Consider the potential $\psi(x,y) = \frac12(x^2+y^2)$ and the rotation ${\bf R}(x,y) = [\cos(y),-\sin(x)]^\top$, with the noise intensity $\sigma^2 = 2$.
We choose a larger number of datasets $M=80$ and increase the number of epochs to 50,000.

\begin{figure}[H]
\begin{center}
\includegraphics[width=0.27\textwidth]{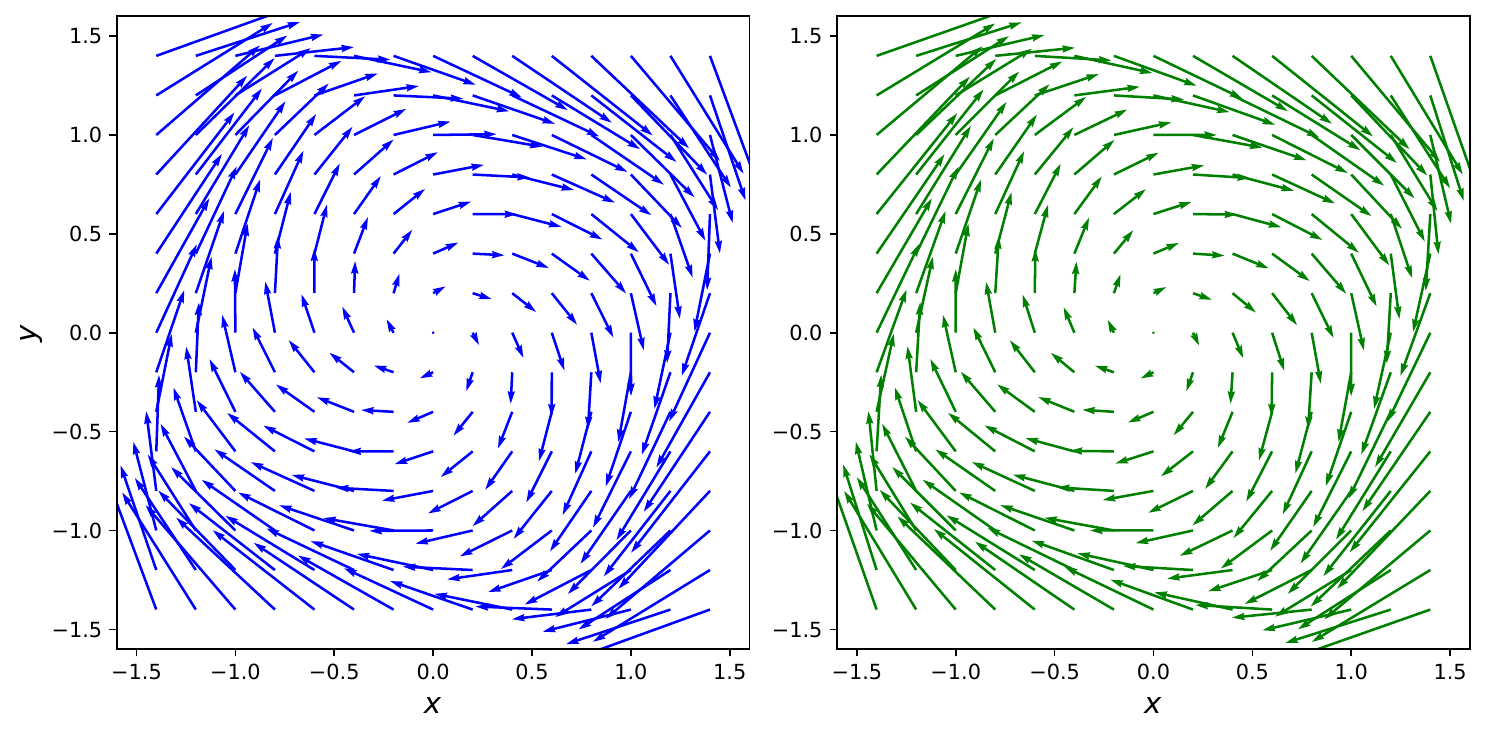}
\includegraphics[width=0.42\textwidth]{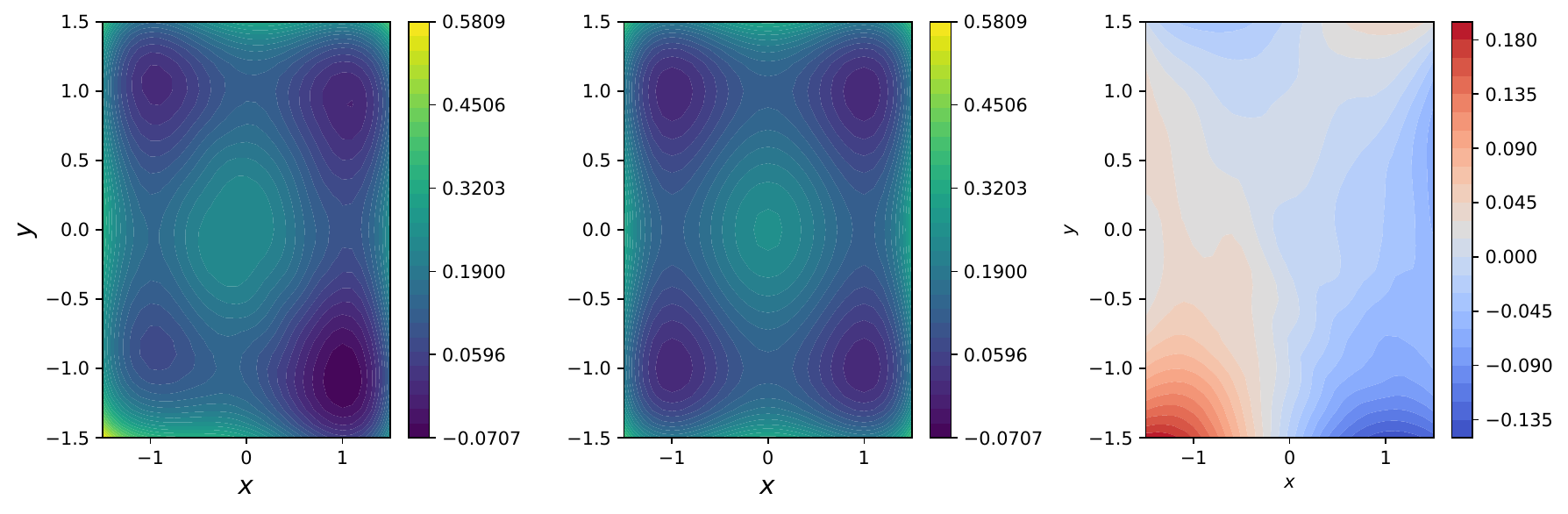}
\includegraphics[width=0.27\textwidth]{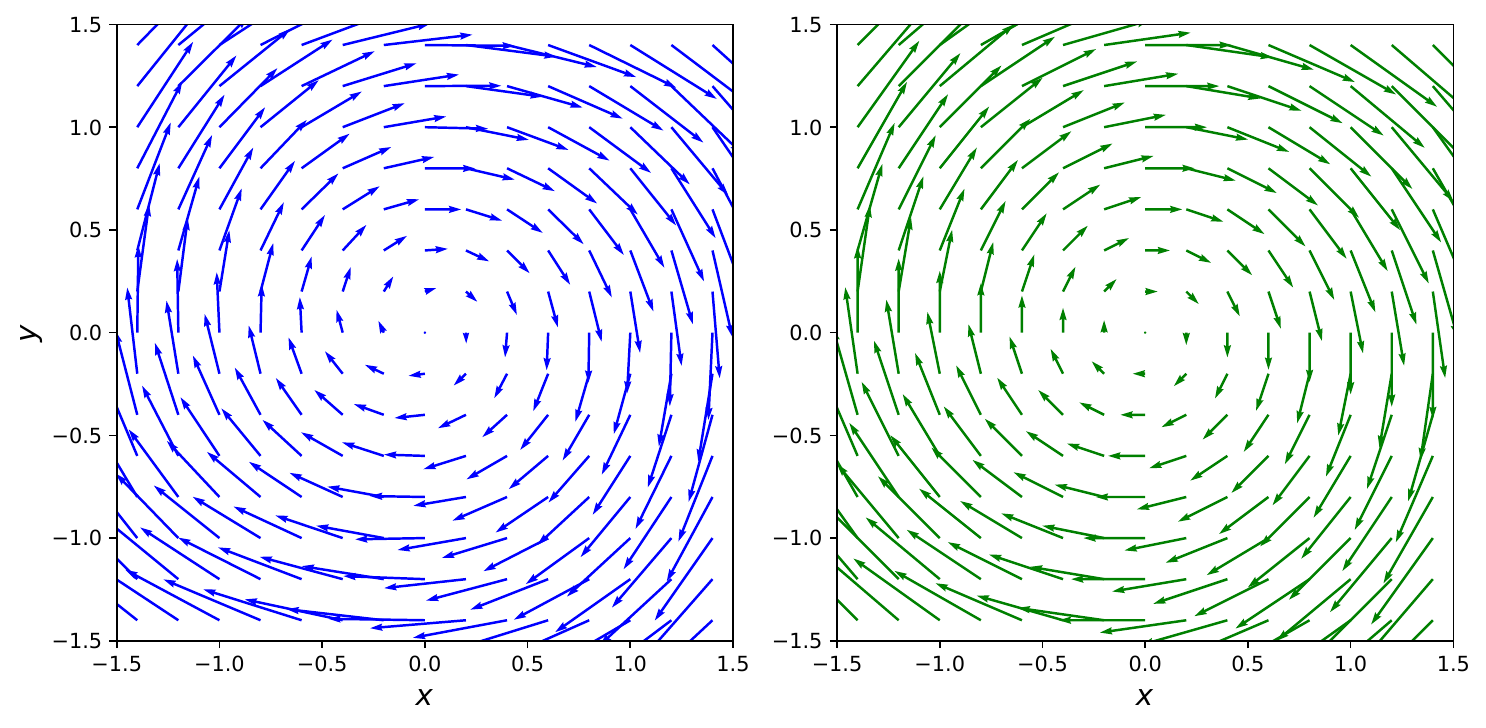} 
\end{center}
\begin{center}
\includegraphics[width=0.27\textwidth]{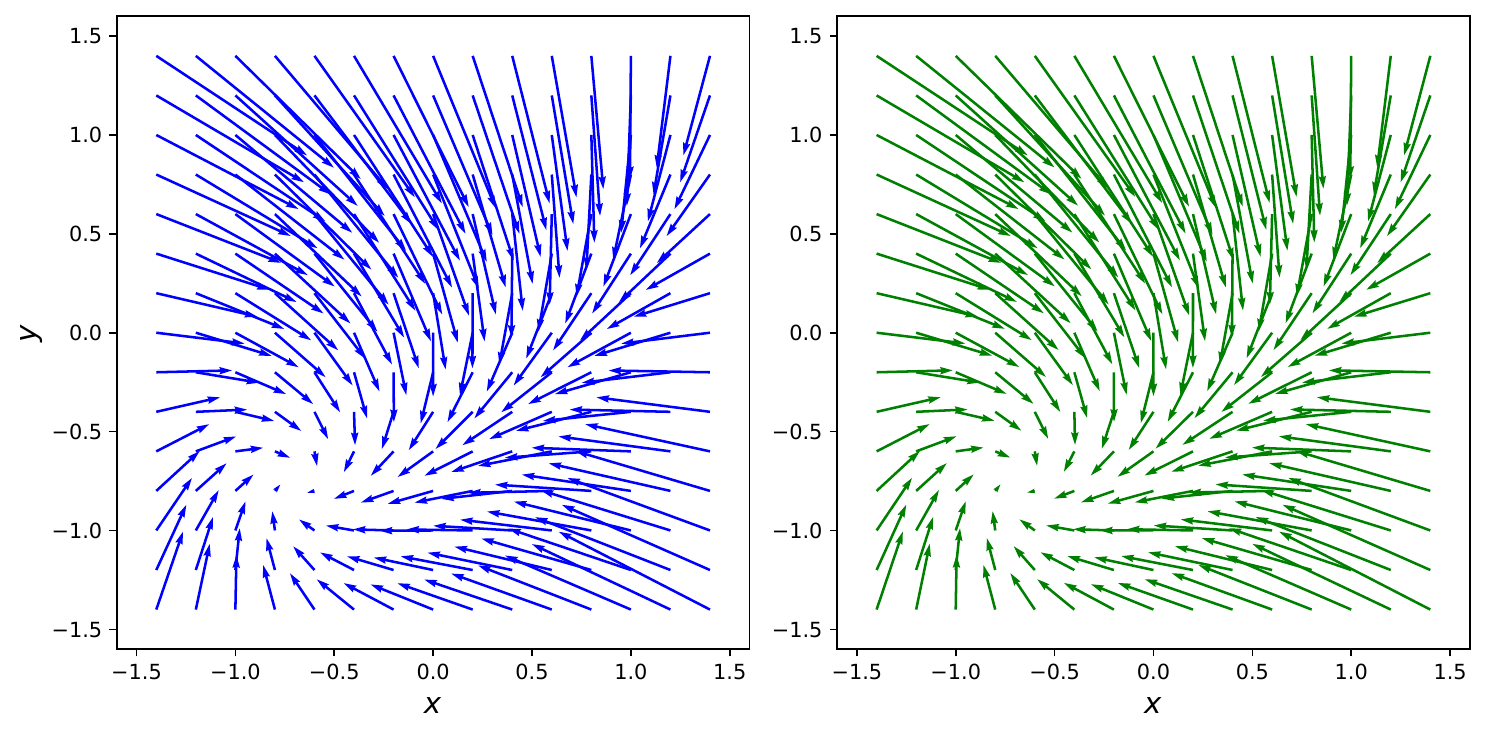}
\includegraphics[width=0.42\textwidth]{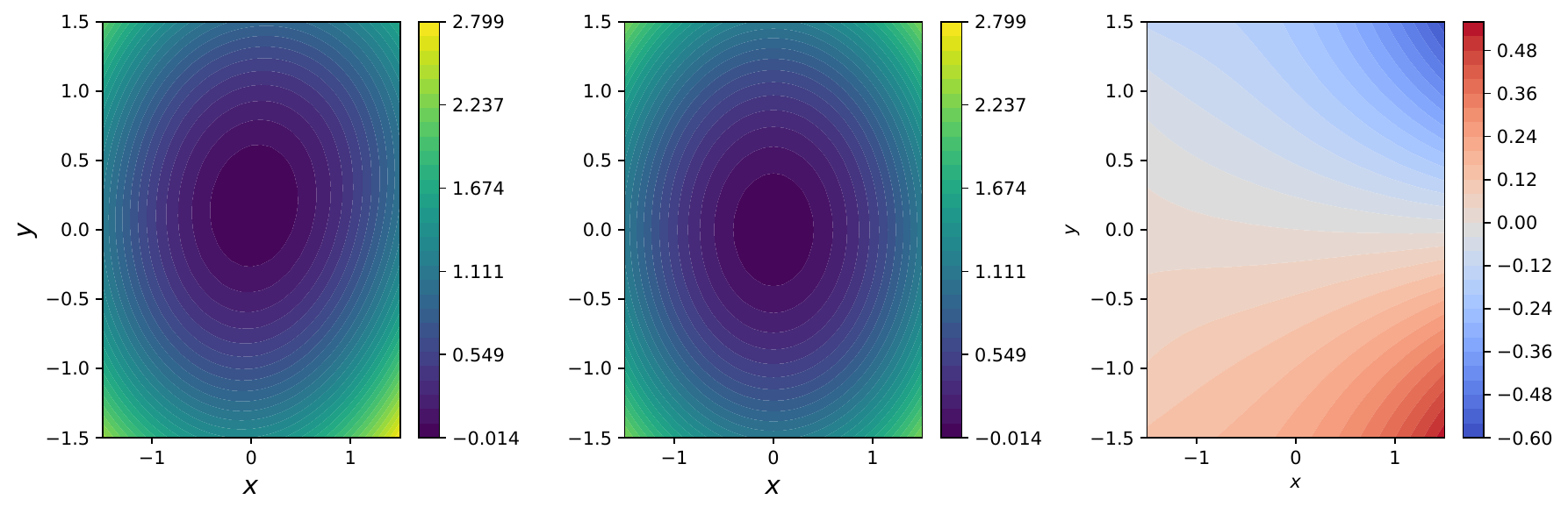}
\includegraphics[width=0.27\textwidth]{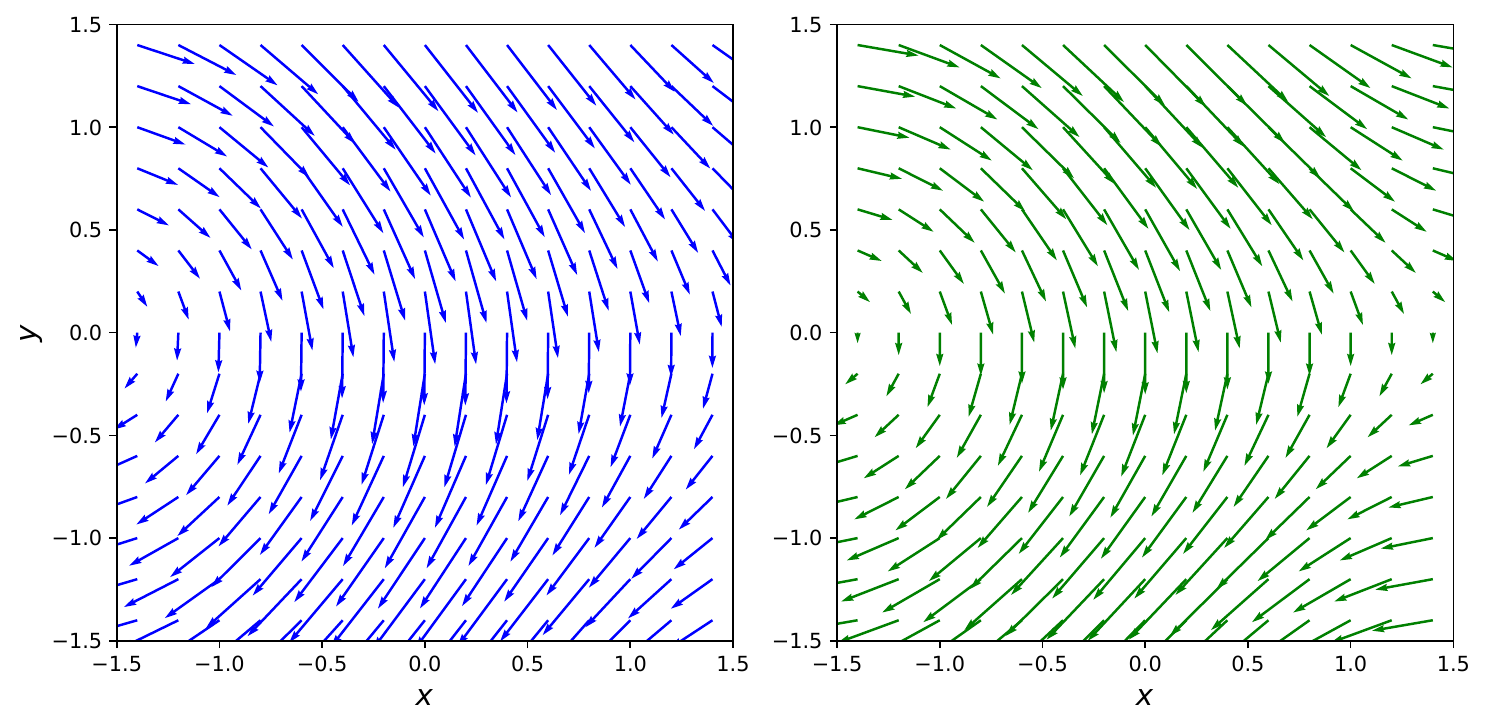} 
\end{center}
  \caption{From left to right: ${\bf b}_{\rm NN}/{\bf b}$ (quiver plots), $\psi_{\rm NN} / \psi / \psi_{\rm NN}-\psi$ (heatmaps), and ${\bf R}_{\rm NN}/{\bf R}$ (quiver plots). {\bf Rows}: (1) Quadruple-well potential, (2) Oscillatory rotation.
  }
  \label{fig:other-examples} 
\end{figure}

For the two examples above, the relative root mean square errors are reported in the first two rows in Table~\ref{tab:other-examples}, and the first two rows in Figure~\ref{fig:other-examples} compare the learned results with the ground truth.

\noindent\textbf{Rough Potential.}
To demonstrate the robustness of our method to rough, oscillatory potentials, we consider $\psi({\bf x}) = \frac14(x^2-1)^2+\frac12y^2+\varepsilon^4\sin(\frac{2\pi x}{\varepsilon})\sin(\frac{2\pi y}{\varepsilon})$ with a parameter $\varepsilon = 0.2$.
Here $\varepsilon$ controls the oscillations: smaller values produce stronger, higher-frequency fluctuations, which grow more pronounced in higher-order derivatives and thus are harder to approximate accurately.
The noise intensity is set to be $\sigma^2 = 2.$

The relative root mean square errors are reported in the last row of Table~\ref{tab:other-examples}.
The last row of Figure~\ref{fig:double-well} compare the learned results with the ground truth.

 {The experimental results summarized in Table \ref{table1} and shown in Figures \ref{fig:other-examples} and \ref{fig:double-well} demonstrate the robustness of the Moment-DeepRitz method under several challenging scenarios. First, the method remains stable in the vanishing noise limit ($\sigma \to 0$), where the pseudo-potential converges to the quasi-potential. Second, it performs reliably on noisy datasets, tolerating perturbations up to a noise level of $0.1$. Finally, it is effective for complex potentials $\psi$ and rotation fields ${\bf R}$ even without imposing the pointwise orthogonality condition between $\nabla \psi$ and ${\bf R}$.
}



\section{Conclusions}
\label{conclusions}

A data-driven approach has been proposed for learning the governing physical laws in complex stochastic dynamics described by the FP equations with non-gradient drifts.  Combining the first-moment dynamics with a variational formulation, we have developed the Moment-DeepRitz method to learn both the pseudo-potential and rotational components in non-equilibrium diffusions.  Compared to the results shown in \cite{KLL2025}, our framework is applicable to the case of non-pointwise orthogonal drift decomposition.  Another advantage of our method lies in its robustness to noisy data and varying noise intensity shown in Section \ref{examples}.  In particular,
we validate the effectiveness through a series of challenging numerical experiments involving quadruple-well potentials, rough potentials and oscillatory rotations.  We remark that the numerical results for the cases  $\sigma^2=10^{-3}$ and $\sigma=0$ in Table \ref{tab:double-well} demonstrate that our framework offers an alternative approach of computing the quasi-potential, parallel to the data-driven method used in \cite{lin2022data}. Numerous open problems remain and a promising research direction in the near future is to extend the data-driven learning framework to the case of time-dependent drifts.




{
\small

 \providecommand{\href}[2]{#2}\begingroup\raggedright\endgroup
}


\appendix

\section{Algorithm for the Moment-DeepRitz Method}\label{sec:algorithm}
Here we present the algorithm for the Moment-DeepRitz method described in Section \ref{framework}.
 \begin{algorithm}
 \caption{Learning generalized diffusions with non-gradient drifts via Moment-DeepRitz method}
 \label{Alg:EnVarA}
 \begin{minipage}{\linewidth}
 \raggedright
 \begin{itemize}
   \item  Choose the training data as $\{(f_j({{\bf x}_{i}},t_1), f_j({{\bf x}_{i}},t_2)\}_{i,j=1}^{N,M},$
  where  $f_j({\bf x}_i,t)$ are probability density functions and $t_1$, $t_2$ are two time steps.
 \item Phase 1: Minimize the loss function (\ref{Ldynamic}) via the deep learning based method, then find the ``best'' parameters of the neural networks to reconstruct ${\bf b}_{\rm NN}$ for learning the drift ${\bf b}$.
     \item Phase 2: Minimize the loss function $I^{\rm dyn}_{\psi}$ given by (\ref{idynloss}) and find the ``best'' parameters of the neural networks to reconstruct ${\psi}_{\rm NN}$ for learning the irrotational component ${\psi}$.
 \end{itemize}
 \end{minipage}
 \end{algorithm}

\section{Results of learning double-well potential under different noise intensities}

\begin{figure}[H]
\begin{center}
\includegraphics[width=0.27\textwidth]{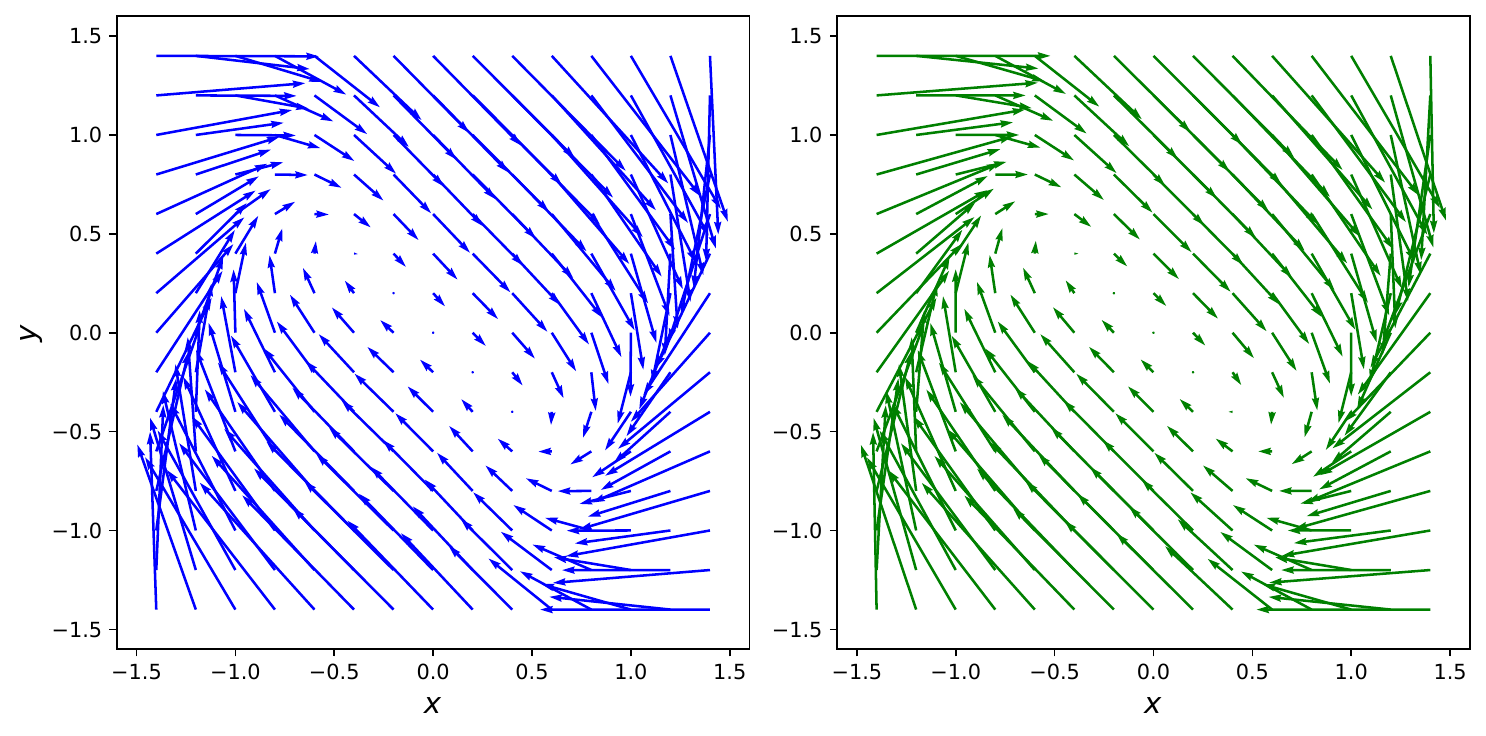}
\includegraphics[width=0.42\textwidth]{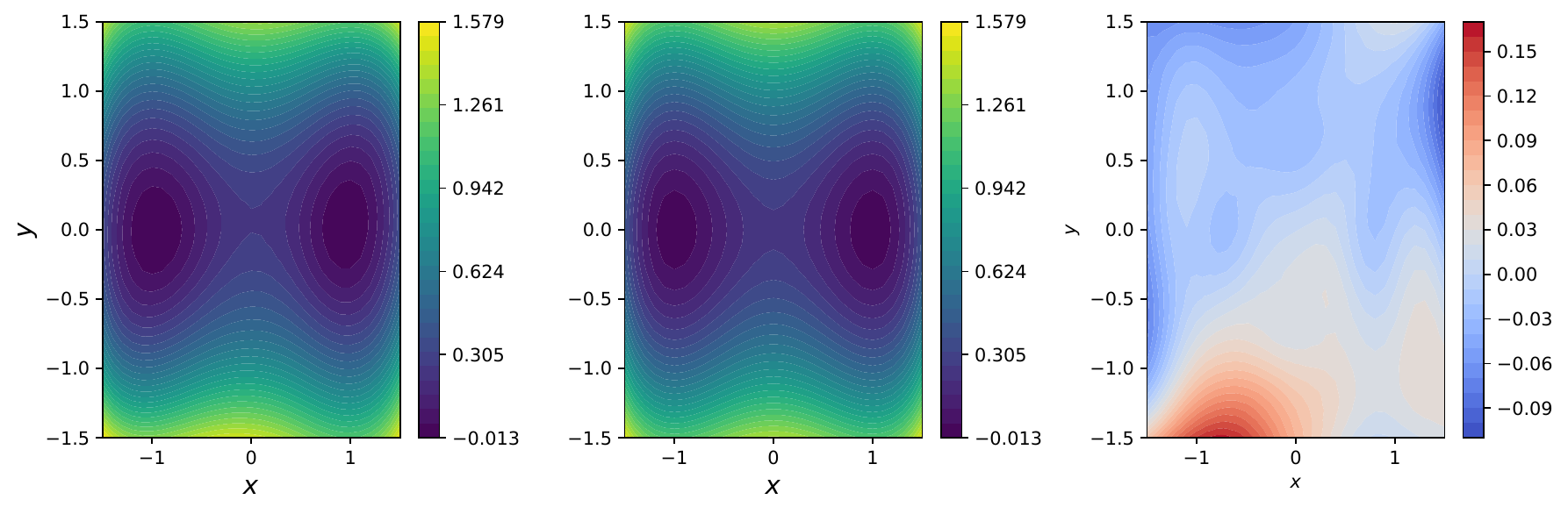}
\includegraphics[width=0.27\textwidth]{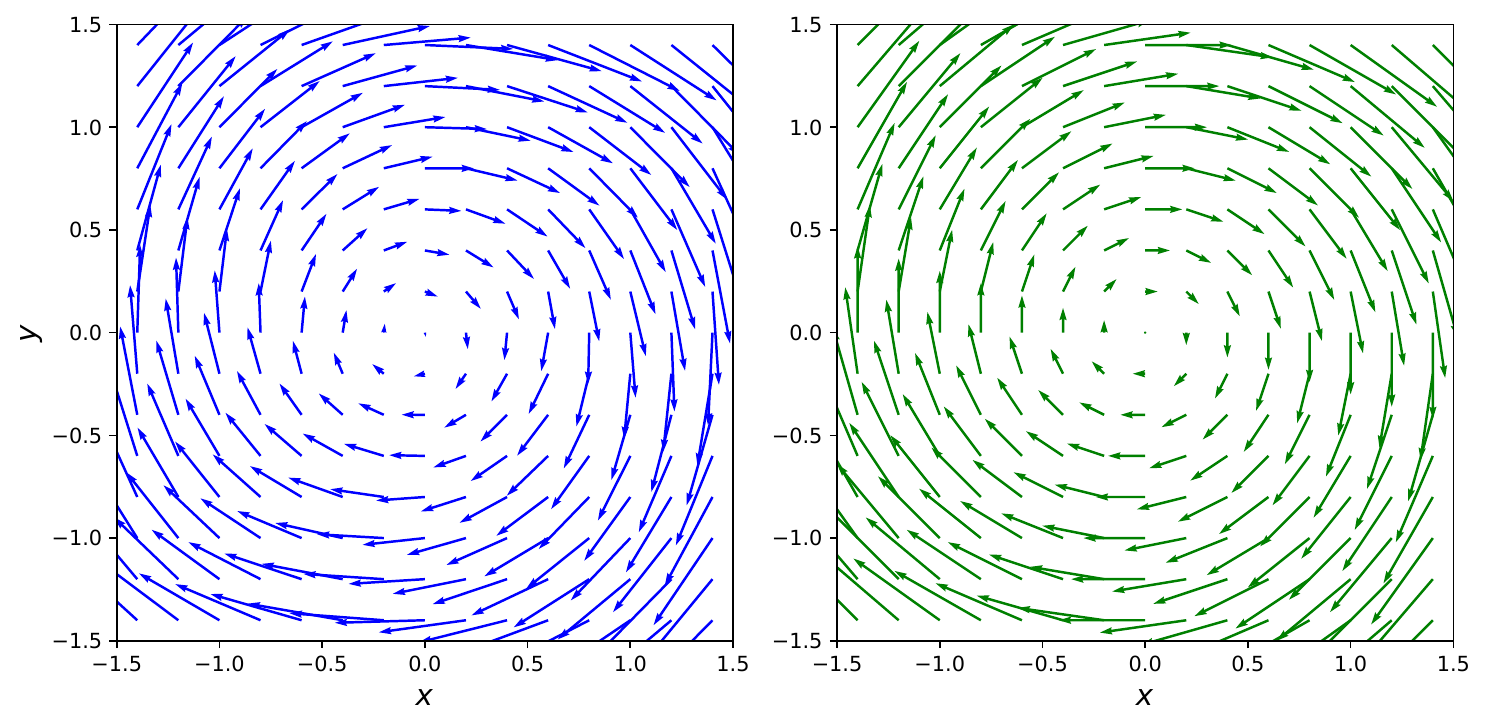} 
\end{center}
\begin{center}
\includegraphics[width=0.27\textwidth]{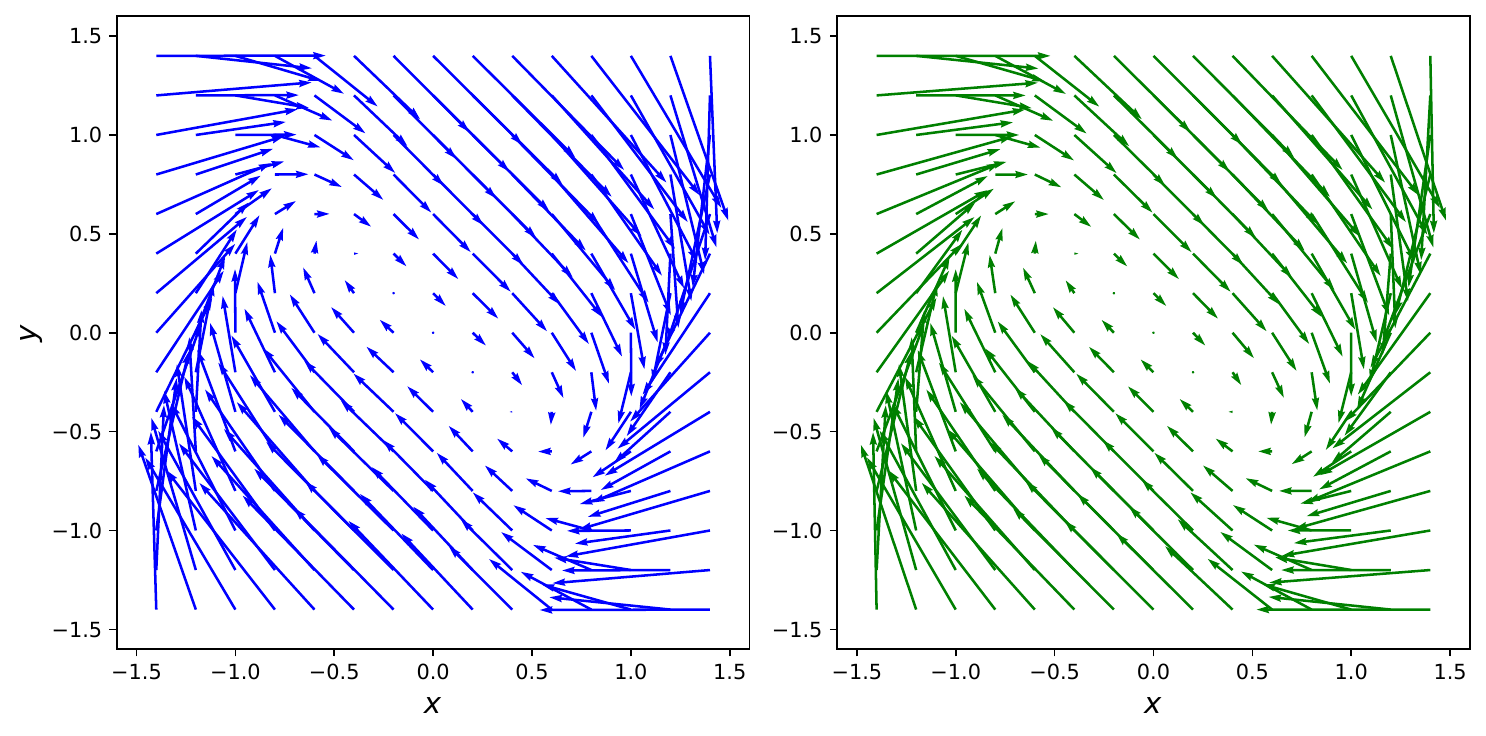}
\includegraphics[width=0.42\textwidth]{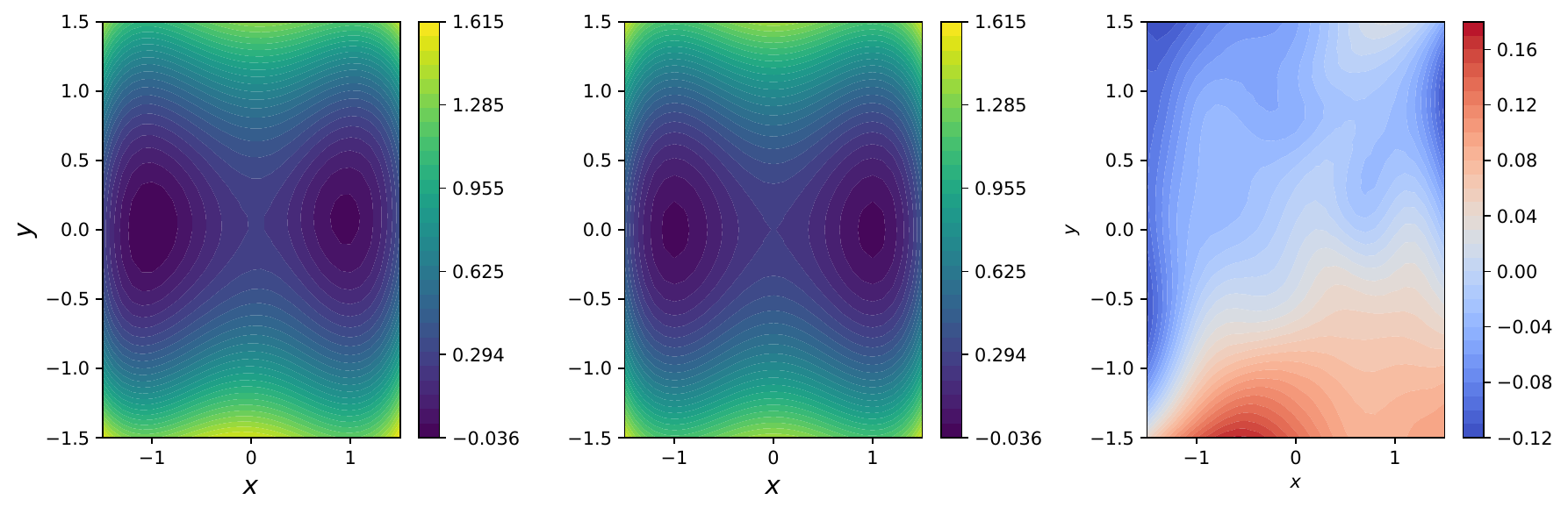}
\includegraphics[width=0.27\textwidth]{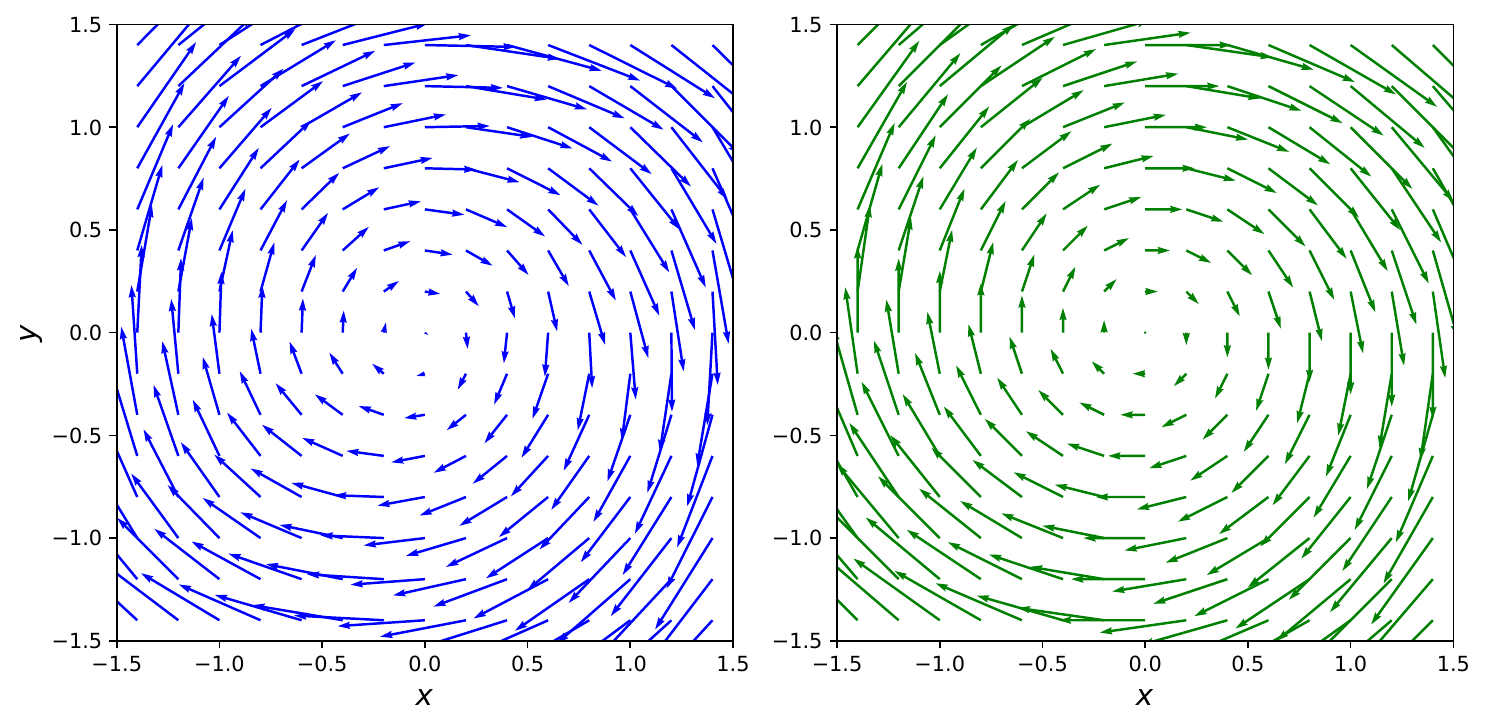} 
\end{center}
\begin{center}
\includegraphics[width=0.27\textwidth]{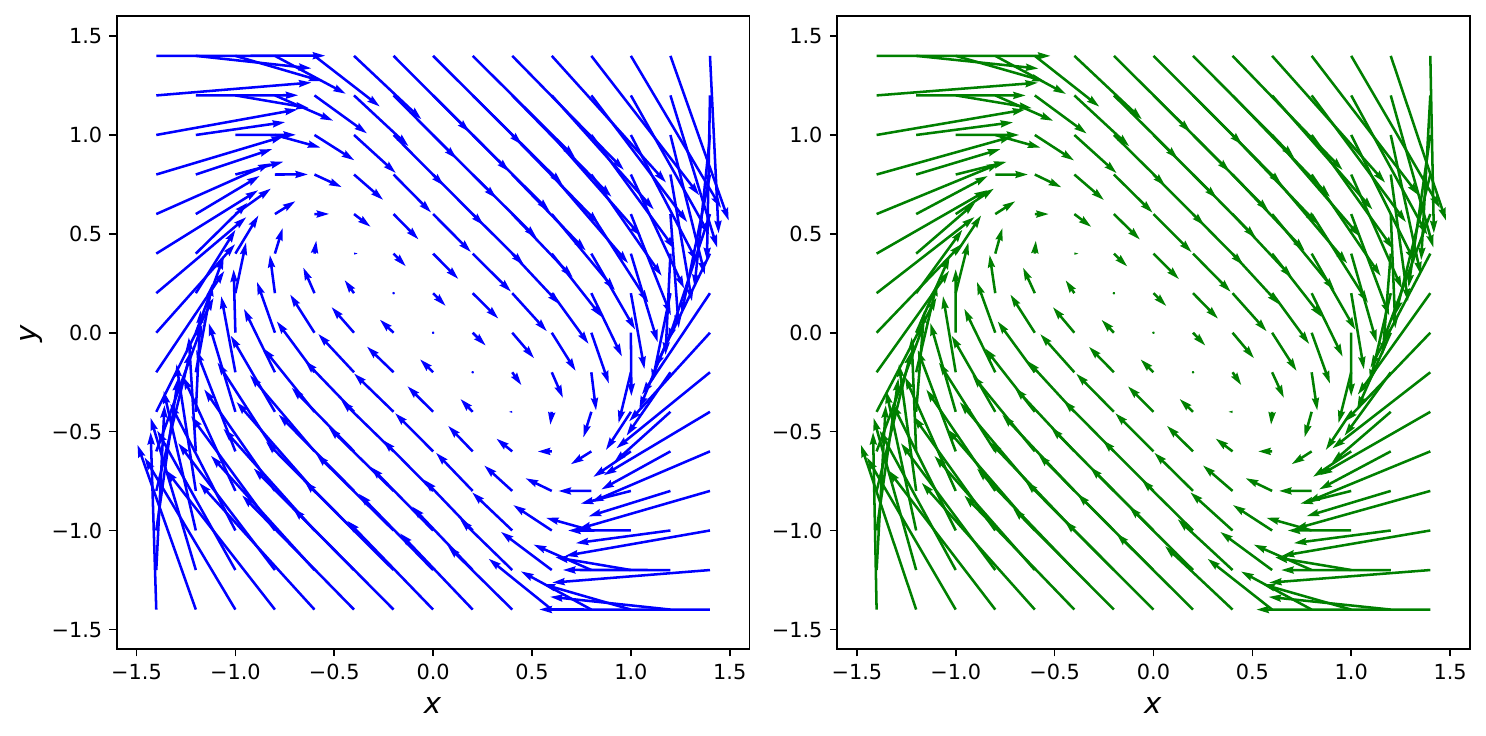}
\includegraphics[width=0.42\textwidth]{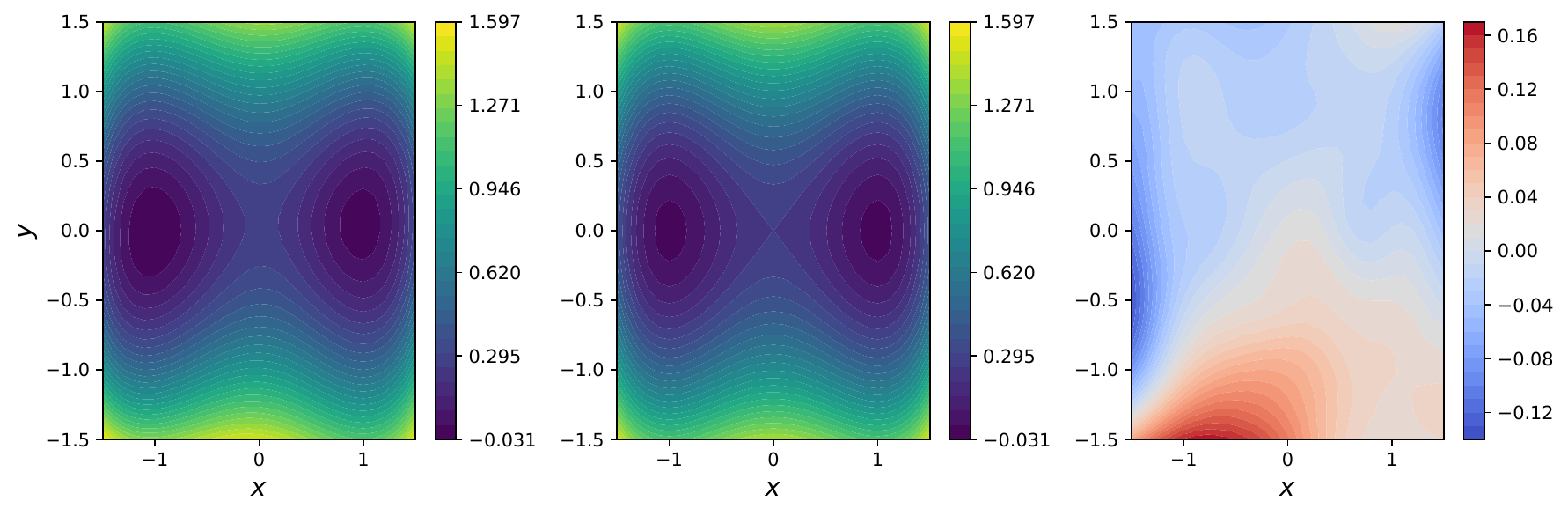}
\includegraphics[width=0.27\textwidth]{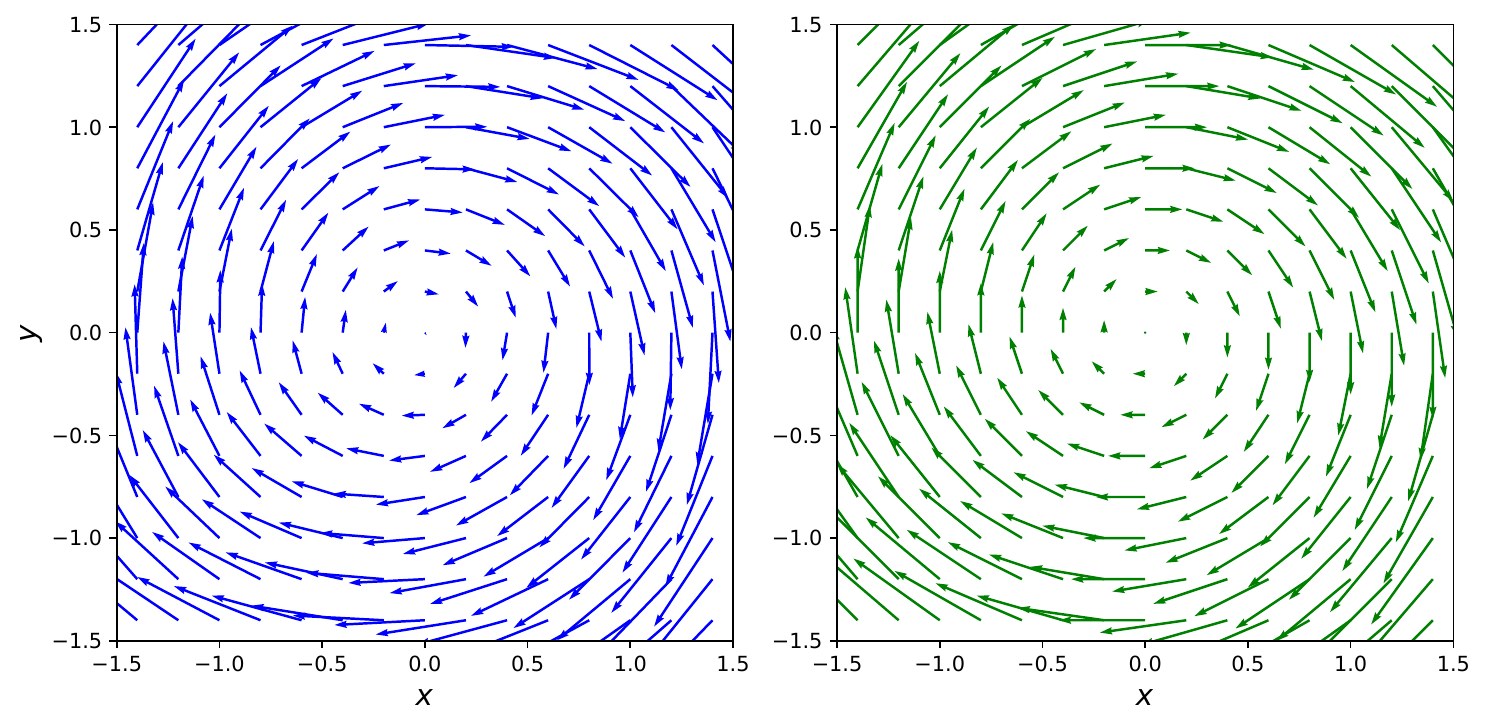} 
\end{center}
\begin{center}
\includegraphics[width=0.27\textwidth]{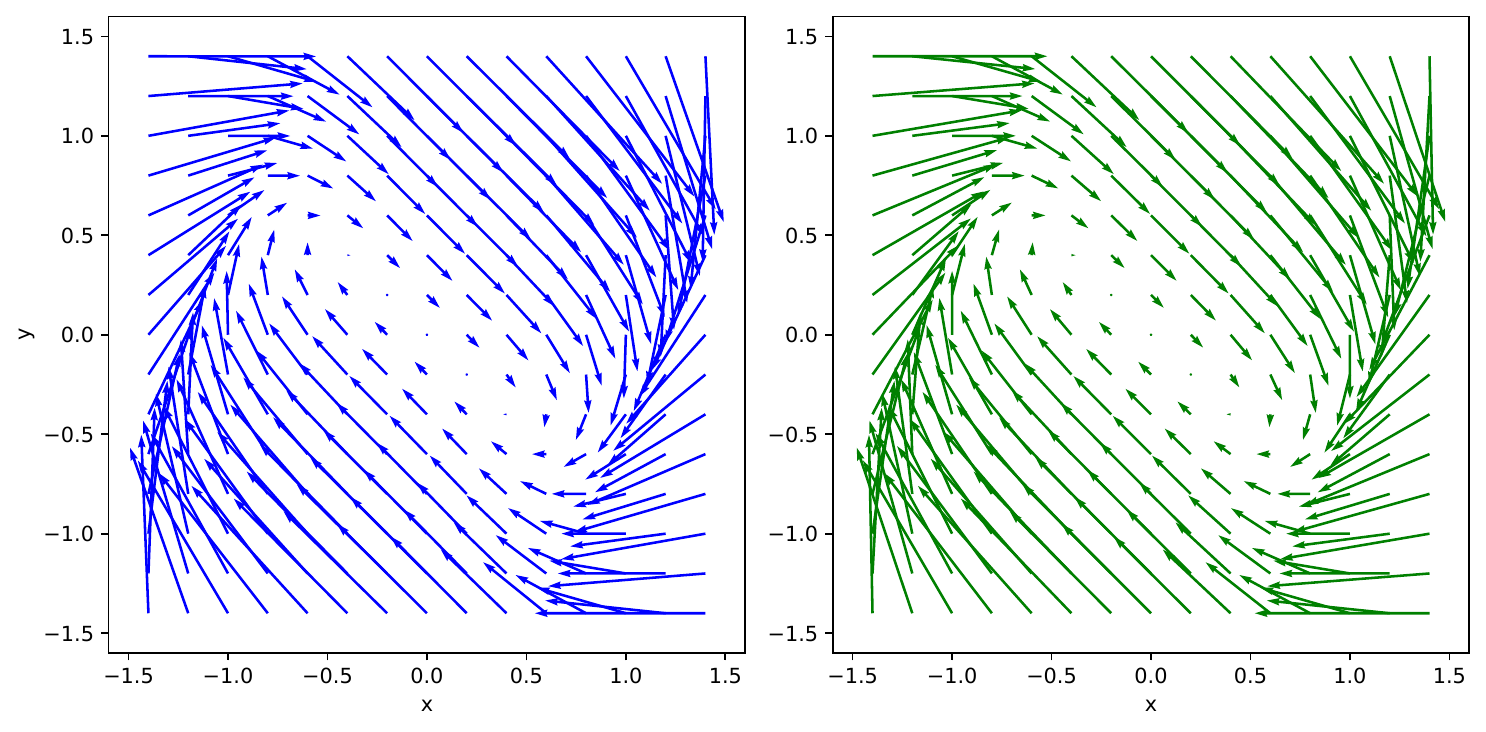}
\includegraphics[width=0.42\textwidth]{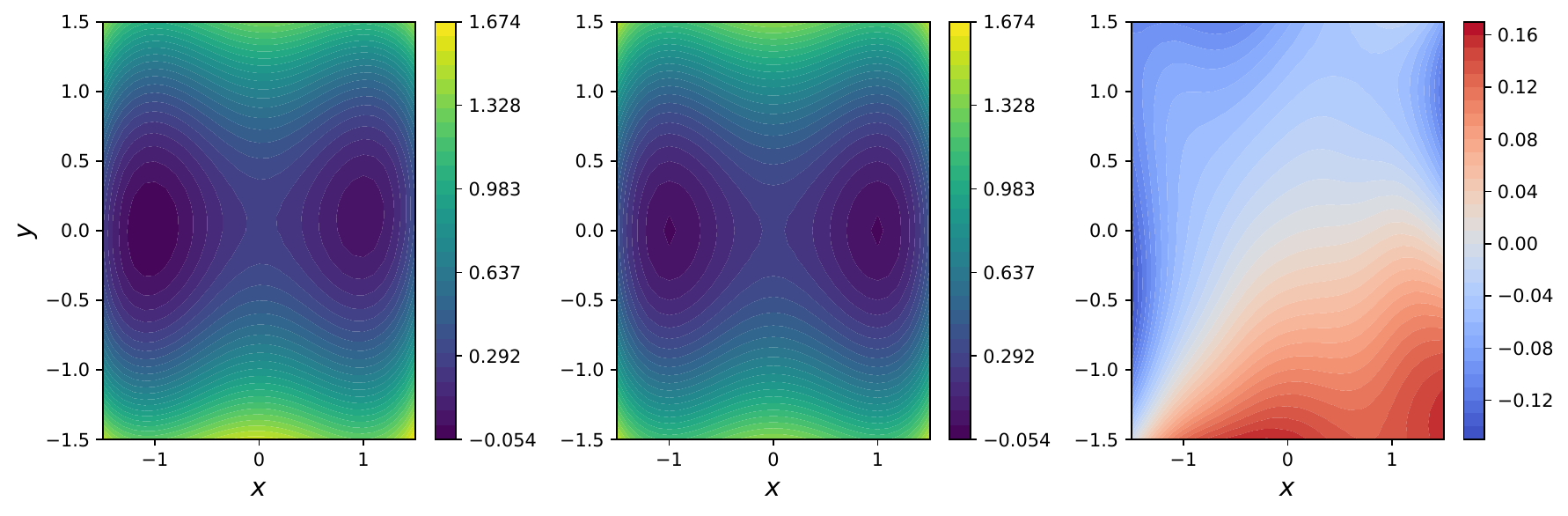}
\includegraphics[width=0.27\textwidth]{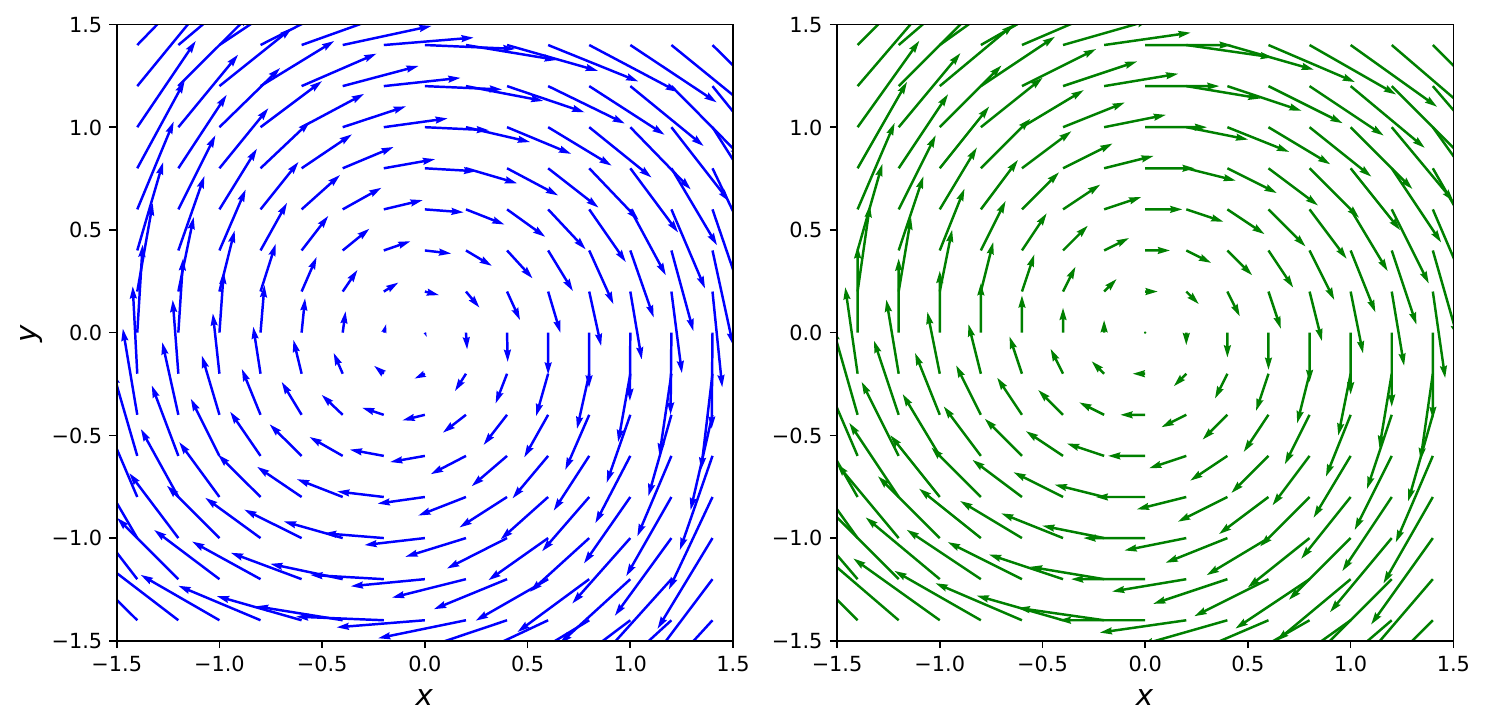} 
\end{center}
  \caption{From left to right: ${\bf b}_{\rm NN}/{\bf b}$ (quiver plots), $\psi_{\rm NN} / \psi / \psi_{\rm NN}-\psi$ (heatmaps), and ${\bf R}_{\rm NN}/{\bf R}$ (quiver plots). {\bf Rows}: (1) $\sigma^2=2$, (2) $\sigma^2=0.001$, (3) $\sigma^2=0$, (4) $\sigma^2=2$ with the rough potential.
  }
  \label{fig:double-well} 
\end{figure}


\newpage
 
\end{document}